\documentclass{article}
\usepackage{ijcai25}

\usepackage{multirow}
\usepackage{graphicx}
\usepackage{amsmath}
\usepackage{amssymb}
\usepackage{booktabs}
\usepackage{bbm}
\usepackage[table,xcdraw]{xcolor}

\usepackage{times}
\usepackage{soul}
\usepackage{url}

\usepackage[pagebackref,breaklinks,colorlinks]{hyperref}
\usepackage{cleveref}

\usepackage[utf8]{inputenc}
\usepackage[small]{caption}
\usepackage{graphicx}
\usepackage{amsmath}
\usepackage{amsthm}
\usepackage{booktabs}
\usepackage{algorithm}
\usepackage{algorithmic}
\usepackage[switch]{lineno}

\usepackage{bbding}

\title{Learning Real Facial Concepts for Independent Deepfake Detection}

\author{
Ming-Hui Liu$^1$ \and
Harry Cheng$^2$ \and
Tianyi Wang$^3$ \and
Xin Luo$^1$ \and
Xin-Shun Xu$^{1*}$
\affiliations
$^1$School of Software, Shandong University\\
$^2$School of Computing, National University of Singapore\\
$^3$Nanyang Technological University\\
\emails
liuminghui@mail.sdu.edu.cn,
xaCheng1996@gmail.com,
terry.ai.wang@gmail.com,
luoxin.lxin@gmail.com,
xuxinshun@sdu.edu.cn
}

\newcommand{\etal}{\textit{et al.}}
\newcommand{\ie}{\textit{i.e.}}

\begin{document}

\maketitle
\begin{abstract}
Deepfake detection models often struggle with generalization to unseen datasets, manifesting as misclassifying real instances as fake in target domains. This is primarily due to an overreliance on forgery artifacts and a limited understanding of real faces. To address this challenge, we propose a novel approach $\mathrm{RealID}$ to enhance generalization by learning a comprehensive concept of real faces while assessing the probabilities of belonging to the real and fake classes independently. $\mathrm{RealID}$ comprises two key modules: the Real Concept Capture Module ($\mathrm{RealC}^2$) and the Independent Dual-Decision Classifier ($\mathrm{IDC}$). 
With the assistance of a Multi-Real Memory, $\mathrm{RealC}^2$ maintains various prototypes for real faces, allowing the model to capture a comprehensive concept of real class. Meanwhile, $\mathrm{IDC}$ redefines the classification strategy by making independent decisions based on the concept of the real class and the presence of forgery artifacts. Through the combined effect of the above modules, the influence of forgery-irrelevant patterns is alleviated, and extensive experiments on five widely used datasets demonstrate that $\mathrm{RealID}$ significantly outperforms existing state-of-the-art methods, achieving a $1.74\%$ improvement in average accuracy. 
\end{abstract}    
\section{Introduction}
\label{sec:intro}
\begin{figure}[t]
\centering
\includegraphics[width=8.6cm]{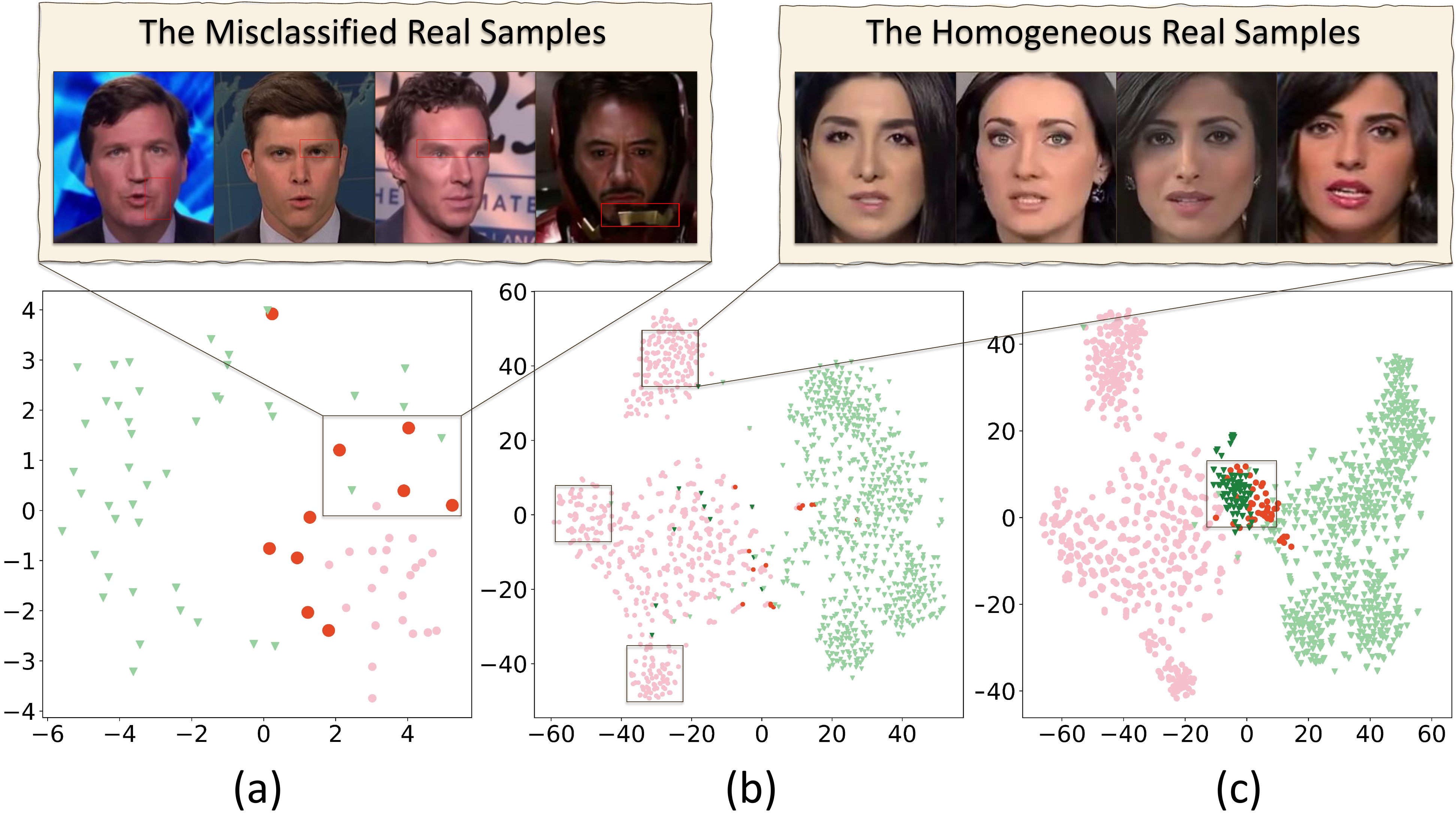}
\setlength{\belowcaptionskip}{-0.3cm} 
\setlength{\abovecaptionskip}{-0.2cm} 
\caption{(a) The real samples are misclassified as fake (highlighted in red), due to the local imperfections that are not actual forgery traces; (b) The features of real faces tend to cluster more tightly; (c) Non-overlapping test samples leading to potential misclassification.}
\label{fig:tsne1}
\end{figure}

With the rapid development of generative algorithms, the barrier to the synthesis of facial forgeries has decreased significantly, fostering online fraud, opinion manipulation, and pornographic content dissemination. To mitigate the abuse of face forgeries, a considerable number of effective deepfake detection methods~\cite{xia2024advancing,sun2024rethinking} have been proposed, and they have achieved significant success when training and testing on the same dataset, \ie, evaluating under the within-dataset setting. Most of these methods treat deepfake detection as a binary classification problem, aiming to capture various artifacts in facial images, such as inter-frame correlation~\cite{1} and style latent flow~\cite{2} to distinguish between real and fake. However, when these models are applied to unseen datasets, their performance often degrades significantly. This issue, \ie, the lack of generalization, has garnered widespread attention in recent years. The solutions for it can be broadly categorized into two approaches: 1) Expanding the scale or diversity of the training dataset to include more complex forgery traces; 2) Optimizing the model architecture to extract general forgery artifacts that are consistent across various deepfake techniques.
Based on the aforementioned approaches, many studies achieve remarkable detection performance. However, as illustrated in Figure \ref{fig:tsne1}(a), the existing models counterintuitively misclassify real instances as fake ones, rather than maintaining a balanced misclassification rate between real and fake instances~(red circles and green triangles represent real and fake samples, respectively).

We attribute this phenomenon to two key factors: 
1) \textbf{Data}. As shown in Figure~\ref{fig:tsne1}(b), real facial features exhibit a clustered distribution, while the distribution of fake ones is relatively uniform. This will cause the model to overfit these homogeneous real samples and limit the feature distribution of the real class. Meanwhile, in Figure~\ref{fig:tsne1}(c), we find test samples that shift beyond the original training distribution are likely to be misclassified. This indicates that the narrow distribution of the real class makes it difficult to cover the real samples in the test set and classify the extra-distributed real samples accurately.
2) \textbf{Classifier}. Deepfake detectors rely on basic binary classifiers to differentiate between real and fake instances. However, because of the difficulty in obtaining the concept of real faces, the classifier would overly rely on forgery traces to make judgments. In other words, the decision-making process has a built-in priority: the model could first determine whether an image is fake according to forgery artifacts, and only if it is not classified as fake will it be recognized as real. Moreover, due to the limited diversity of the training dataset, forgery-irrelevant patterns, such as noise and blur, will also be entangled with forgery artifacts. When these so-called `mis-artifacts' appear in real samples during the testing phase, they can easily lead to an incorrect judgment. 
Given the insufficient exploration of real samples and the overlook of real facial concepts, a natural question motivates us -- Could we improve the generalization by learning a more comprehensive concept for real faces and inducing the model to make judgments based on both the real facial concepts and the presence of forgery artifacts?

We address the aforementioned issues by proposing the Real Facial Concepts based Independent Deepfake Detection~(RealID) approach. Our approach focuses on exploring and leveraging real facial concepts, enabling the model to make judgments based on both real facial concepts and forgery artifacts. Specifically, we design two novel modules:
1)~Real Concept Capture Module~(RealC$^2$), which employs a Multi-Real Memory mechanism to store various real prototypes. It operates over the entire training set, learning a more comprehensive real facial concept with specially designed Prototype Distinction Loss and Prototype Diversity Loss. In this way, the model can focus more on the differences of the real samples and generate a more robust distribution for the real class.
2)~Independent Dual-Decision Classification strategy~(IDC). We reformulate the binary classification strategy to a novel independent dual-decision strategy. By adding a regularization term and two auxiliary classes, the probability of belonging to the incorrect classes can further decrease after being optimized with the cross-entropy loss. This means that our IDC can adaptively search for an alternative optimization path, \ie, utilizing the real facial concepts learned from RealC$^2$, and mitigate the misguidance of the overfitting mis-artifacts. Using previously overlooked real facial concepts, our method enables the model to make robust decisions from both real and fake perspectives, thereby enhancing its generalization capability in target domains. We conduct extensive experiments on five widely used datasets and the experimental results demonstrate a $1.74\%$ average accuracy improvement compared to existing state-of-the-art methods. 

In summary, our contributions are three-fold:
\begin{itemize}
    \item We attribute the issue of limited model generalization to the inadequacy in learning comprehensive real facial concepts. Our observations indicate that existing detectors tend to overly rely on forgery artifacts, which often results in the misclassification of real instances as fake. 
    \item We propose the Real Facial Concepts based Independent Deepfake Detection approach, which combines two special modules to facilitate the learning of real facial concepts and enable the classifier to independently assess the probabilities of belonging to real and fake classes.
    \item Extensive experiments are conducted to demonstrate the effectiveness of our approach which achieves state-of-the-art generalization performance on several datasets.
\end{itemize}
\section{Related Work}
\begin{figure*}[t]
{\includegraphics[width=1\textwidth]{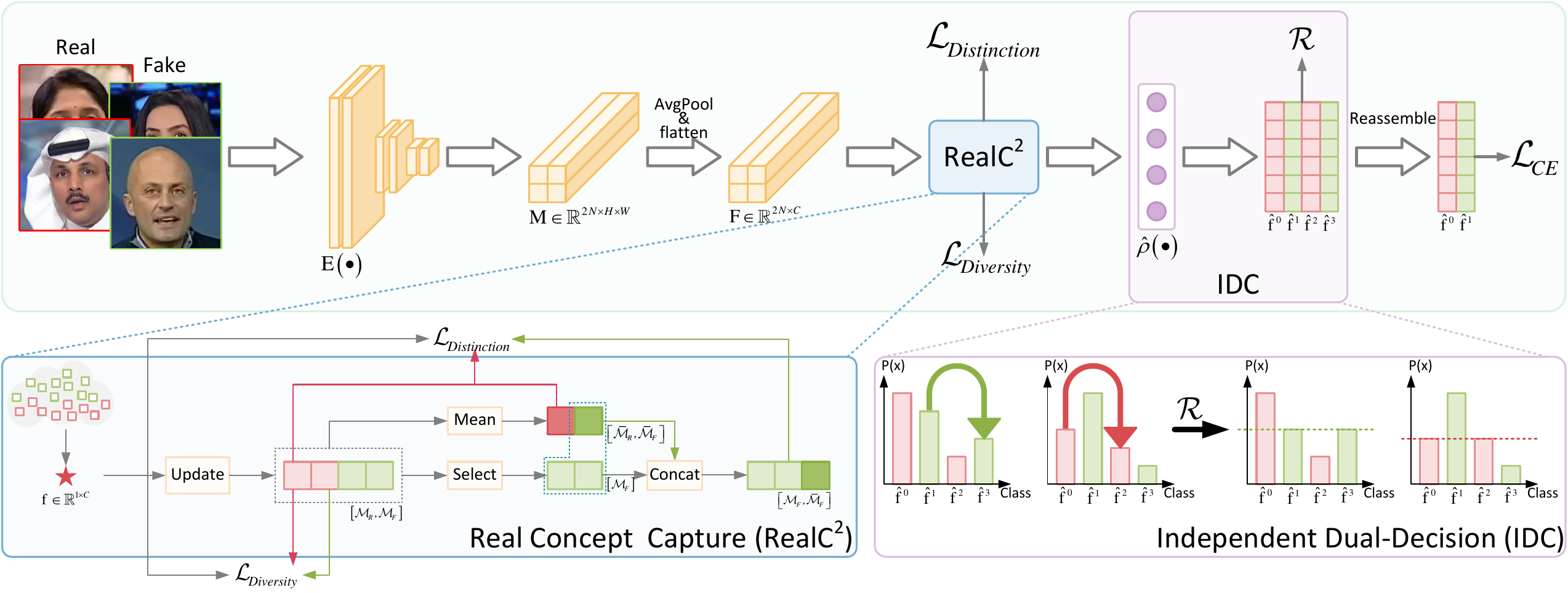}}
\setlength{\abovecaptionskip}{-0.1cm} 
\setlength{\belowcaptionskip}{-0.1cm}
\caption{Overall architecture of the RealID framework. RealID consists of two main modules: (i) the Real Concept Capture($\mathrm{RealC^2}$) module, which uses a multi-real memory mechanism to learn a more comprehensive real facial concept, and (ii) the Independent Dual-Decision(IDC) module, which leverages regularization terms to independently optimize the decision-making process for different categories.}
\label{fig:RealID}
\end{figure*}
\subsection{Deepfake Generation}
The rapid development of portrait synthesis has propelled deepfake technology~\cite{DF_PAMI4} into a prominent research area. Autoencoders~\cite{autoencoder} form the foundational architecture for many early deepfake approaches~\cite{Face2Face,S_Ob}. Typically, these methods involve training two separate models on a reconstruction task and then swapping their decoders to change the identities of source faces. While these techniques can produce realistic face-swapped images, they are limited to one-to-one face swapping.
To enable more versatile synthesis, Generative Adversarial Networks (GANs)~\cite{GAN} have become increasingly popular due to their ability to generate arbitrary high-quality facial images. For example, StyleGAN~\cite{StyleGAN} allows for the manipulation of high-level facial attributes by using a progressively growing structure combined with adaptive instance normalization. IPGAN~\cite{IPGAN}, on the other hand, disentangles the identity and attributes of the source and target faces, which are then blended to synthesize a new face.
Recently, identity-relevant features have been integrated into deepfake generation to enhance identity consistency. Xu~\etal~\cite{Region_aware_swapping} refined identity consistency by augmenting both local and global identity-relevant features through cross-scale semantic interaction modeling, achieving more coherent face swapping that maintains identity integrity.

\subsection{Deepfake Detection}
Deepfake detection~\cite{10.1145/3699710,Hong_Deepfake_CVPR_2024,CVPR24_Yan_Aug,xia2024mmnet,guan2024improving} is generally cast as a binary classification task. Preliminary efforts often endeavor to detect the specific manipulation traces~\cite{Exploring_Frequency_Adversarial}. Masi~\textit{et~al.}~\cite{Two-Branch} utilized optical and frequency artifacts separately. SSTNet~\cite{SSTNET} detects edited faces through spatial, steganalysis, and temporal features. These models have shown certain improvements on some datasets. However, they often encounter inferior performance when applied to different data distributions. Several approaches are proposed to address this issue of lack of generalization~\cite{Tan_CVPR24,li2023logical}. 
One manner is to construct datasets that encompass a wider range of forgery methods~\cite{FakeAVCeleb,Openfor,DiFF}.
For instance, the KoDF dataset~\cite{KODF} comprises over 200,000 videos generated by six distinct algorithms, while DF-Platter~\cite{df-platte} involves over 130,000 multi-face forgery videos. These high-quality datasets encompass diverse data sources. However, it is crucial to note that these datasets often place greater emphasis on the fake side, while the distribution of real images tends to be more homogeneous. This imbalance makes it easier for a model to overfit on real features, which in turn leads to misclassification.
Another approach is to learn more generalized features of forgeries~\cite{RFM,SRM,MAT,x-ray,DF_PAMI1}. For instance, RealForensics~\cite{Leveraging_Real_Talking} exploits the visual and auditory~\cite{VFD} correspondence in real videos. Chen~\etal~\cite{Self_ADV} specified the blending regions and facial attributes to enrich the deepfake dataset with more manipulation types. Shiohara~\etal~\cite{SBI_ShioharaY22} introduce a self-blended framework to capture boundary-fusion features. These methods have achieved notable success in capturing forgery artifacts, reaching performance plateaus on multiple datasets. However, they tend to lack sufficient focus on real instances, leading to an imbalanced decision-making process.

\section{Methodology}
As illustrated in Figure~\ref{fig:RealID}, our RealID framework improves generalization through the combination of two novel modules: the Real Concept Capture~($\mathrm{RealC^2}$) module and the Independent Dual-Decision~($\mathrm{IDC}$) module. Specifically, $\mathrm{RealC^2}$ uses a Multi-Real memory mechanism~\cite{park2020learning} to maintain several prototypes of real faces, enabling the model to extract more comprehensive features from real samples and preventing the model from overfitting to the homogeneous training sets. $\mathrm{IDC}$ modifies the decision logic of the naive deepfake classifier, which makes decisions based solely on the presence or absence of forgery artifacts. By incorporating an additional regularization term and auxiliary classes, it can independently reduce the probability of belonging to the false classes through different optimization paths. This means $\mathrm{IDC}$ mitigates the influence of mis-artifacts and improves the generalization of the deepfake detector.

During the training process, we specifically construct mini-batches to ensure the inclusion of both $N$ real samples ${{\mathbf X}^{\rm real}}$(with the labels of ${\mathbf Y=0}$) and $N$ fake samples 
${{\mathbf X}^{\rm fake}}$(with the labels of ${\mathbf Y=1}$). Then, we obtain the feature maps ${{\mathbf M} \in {\mathbb{R}^{2N \times H \times W}}}$ via a feature extractor $\mathrm E(\cdot)$ and transform them to feature vectors ${{\mathbf F} \in {\mathbb{R}^{2N \times C}}}$ by a series of pooling and flattening operations. This feature extraction process can be formalized as follows:
\begin{equation}
\begin{split}
{\mathbf{F}} = \mathrm{Flatten\left( {AvgPool\left( {E\left( {\mathbf{X}}\right)} \right)} \right)}.
\end{split}
\end{equation}

\subsection{Real Concept Capture Module} 
As shown in Figure~\ref{fig:tsne1}(b), the homogeneous real samples in the training set may mislead the detection models into overfitting to a specific cluster while neglecting the concept of real faces. This limitation makes it difficult to address the distribution shift in the testing set. To overcome this issue, we design the $\mathrm{RealC^2}$ module to extract more comprehensive real facial features. In particular, instead of treating the real class as a single homogeneous entity, our model subdivides real facial features with the assistance of diverse real prototypes. In this way, the model can distinguish the fine-grained differences between real instances and acquire a more comprehensive understanding of real faces.

\noindent\textbf{Prototype Initialization.} 
Before training begins, we randomly initialize $K$ vectors ${{\cal M}_R} = \{{\mathbf{m}_1},{\mathbf{m}_2}, \cdots, {\mathbf{m}_K}\}$ to construct the Multi-Real Memory, and each ${\mathbf{m}_i} \in \mathcal{M}_R $ can be considered as a real facial prototype. During the training, these prototypes are designed to record real facial patterns and explicitly enforced to maintain diversity. This prevents the model from focusing solely on the commonalities of real faces (specific implementation will be discussed below).

\begin{figure}[t]
\centering
\includegraphics[width=8.5cm]{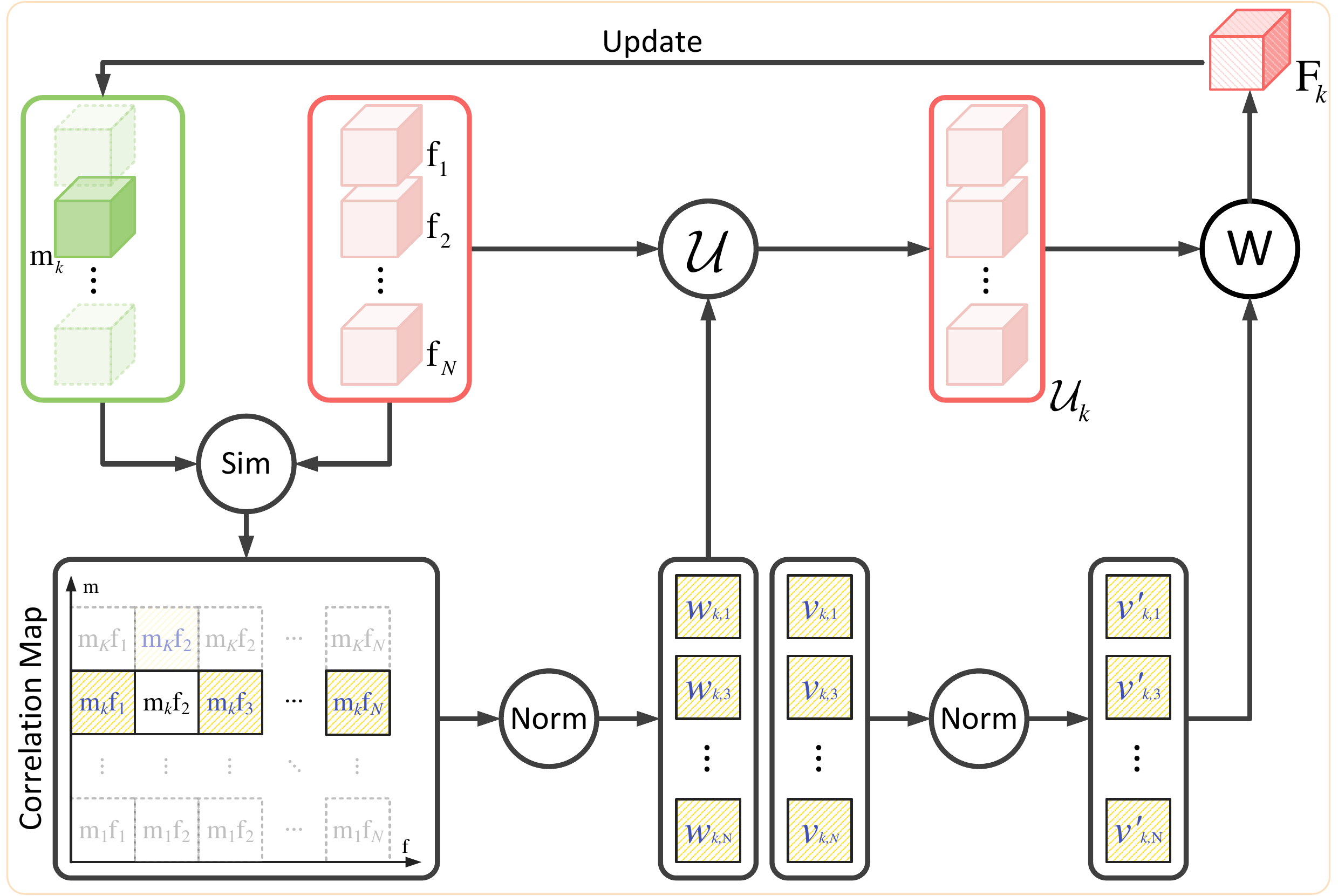}
\caption{Illustration of the update process for real facial prototypes.}
\label{fig:2}
\end{figure}

\noindent\textbf{Prototype Updating.}
In each training iteration, we update the real facial prototypes ${{\cal M}_R}$ with total real facial features in the mini-batch. Specifically, we first calculate the similarity between each prototype $\mathbf{m}_i$ and specific real facial features, resulting in a correlation map of size $K \times N$. Then, we normalize the correlation map using Softmax function along both horizontal and vertical directions. For the vertical one, we perform feature normalization among the $K$ prototypes:
\begin{equation}
{w_{i,j}} 
= \frac{\exp \left( {{\mathbf{m}_i}^T}{\mathbf{f}_j} \right)}
{{\sum\nolimits_{i=1}^K  {\exp \left( {{{{{\mathbf{m} _i}}}^T}{{\mathbf{f}}_j}} \right)} }},
\label{eqn:ver_cos}
\end{equation}
where $\mathbf{f}_j$ is the feature of a specific real instance. Similarly, the horizontal one is performed on the $N$ real features:
\begin{equation}
{v_{i,j}} 
= \frac{\exp \left( {{\mathbf{m}_i}^T}{\mathbf{f}_j} \right)}
{{\sum\nolimits_{j=1}^N  {\exp \left( {{{{{\mathbf{m} _i}}}^T}{{\mathbf{f}} _j}} \right)} }}.
\label{eqn:hor_cos}
\end{equation}
The results in Equation~(\ref{eqn:ver_cos}) and (\ref{eqn:hor_cos}), \ie, $w_{i,j}$ and $v_{i,j}$, represent the vertical and horizontal normalized matching probabilities and guide the selection of the most suitable real facial features for prototypes update. 
Firstly, each sample $\mathbf{f}_j$ is assigned to its nearest prototype $\mathbf{m}_{p_1}$. And we select the prototype $\mathbf{m}_{p_1}$ corresponding to the highest  vertical normalized matching probabilities between $\{ {{w_{1,j}},{w_{2,j}}, \cdots ,{w_{K,j}} }\}$:
\begin{equation}
{p_1} = {\rm argmax}{w_{i,j}}.
\end{equation}
This allocation process means that each sample has exactly one nearest prototype, while the number of samples assigned to each prototype is not fixed. Then, all samples assigned to the same prototype $\mathbf{m}_k$ can be considered as an update set ${{\cal U}_{k}}$. Based on corresponding horizontal normalized matching probabilities $\{ {{v_{k,1}},{v_{k,2}}, \cdots ,{v_{k,n}} }\}$, the prototype $\mathbf{m}_k$ will be updated using a weighted sum of update set ${{\cal U}_{k}}$:
\begin{equation}
\begin{split}
&{\mathbf m}_{k} \leftarrow {\rm Normalize} ( {{{\mathbf m}_{k}} + \sum\limits_{{\mathbf{f}_j} \in {{\cal U}_{k}}} {{{v'}_{{k},j}}{\mathbf{f}_j}} } ),
\end{split}
\end{equation}
where ${\rm Normalize(\cdot)}$ represents the L2 norm function, and ${v'}_{{k},j}$ is the variant of horizontal normalized matching probabilities ${{v}_{{k},j}}$, which is further regularized with the max value in its respective update set ${{\cal U}_{k}}$ as follows:
\begin{equation}
{{v'}_{{k},j}} = \frac{{v}_{{k},j}}{{\mathop {\max }\limits_{_{{\mathbf{f}_{j'}} \in {{\cal U}_{k}}}} {v_{k, j'}}}}.
\end{equation}
To have a more clear perspective, we illustrate the computation and update process related to $\mathbf m_k$ in Figure \ref{fig:2}.

\noindent\textbf{Prototypes Distinction Loss and Diversity Loss.} 
We utilize two specially designed prototype losses to achieve precise prototype learning. First, we employ the Prototypes Distinction Loss (${\cal L}_{\mathrm{Distinction}}$) to align the distribution within the subclass. Specifically, we encourage each real facial feature $\mathbf{f}_j$ to move toward its nearest prototype $\mathbf{m}_{p_1}$ and the mean real facial prototype $\bar {\cal M}_R$. To enhance the distinction between the real and fake classes, we introduce a set of fake facial prototypes\footnote{The fake prototypes are updated using the same method as the real ones and relying on the fake instances in the mini-batch.} ${{\cal M}_F}$. Then, we further push the real facial feature $\mathbf{f}_j$ away from the fake prototypes and the mean fake prototype $\bar {\cal M}_F$. This process can be expressed as:
\begin{equation}
\begin{split}
&{{\cal L}_{\mathrm{Distinction}}} = 
- \log \frac{\exp ( {{{\rm m}_{p_1}}^T{{\rm f}_j}} ) + 
\exp ( {{{\bar {\cal M}}_R}^T{{\rm f}_j}} )}
{\sum\limits_{k=K}^{2K}{\exp ({{{\rm m}_k}^T{{\rm f}_j}})} + 
\exp ( {{{\bar {\cal M}}_F}^T{{\rm f}}_j} )}, 
\end{split}
\end{equation}
where the mean prototype $\bar {\cal M}$ can be calculated as follows:
\begin{equation}
\begin{split}
&{{\bar {\cal M}}} = \frac{{\sum\limits_{k=1}^K {{\rm m}_k} }}{K}.
\end{split}
\end{equation}

In addition, due to the lack of explicit supervision for sub-classes, the Prototypes Distinction Loss may mislead different real subclasses to collapse into an undesirable whole. To address this, we design a Prototypes Diversity Loss to maintain the diversity of real facial prototypes:
\begin{equation}
\begin{split}
&{{\cal L}_{\mathrm{Diversity}}} = \sum\limits_{j=1}^N 
{{[{{{\|{{{\mathbf f}_j}-{\mathbf m_{p_1}}}\|}_2}-{{\|{{{\mathbf f}_j}-{\mathbf m_{p_2}}}\|}_2}+\alpha }]}_+} \\
&{p_2} = \mathop{\rm argmax} \limits_{i \ne {p_1}} {w_{i,j}},
\end{split}
\end{equation}
where $\mathbf m_{p_2}$ is the second nearest prototype of ${\mathbf f}_j$. $\alpha$ is a predefined margin. In this way, the features of different sub-classes can maintain a proper distance during training.

\subsection{Independent Dual-Decision Module}
Because of the difficulty in learning and leveraging real facial concepts, deepfake detection models would classify samples with forgery-related features as fake and classify samples without those features as real. 
Traditionally, deepfake detectors incorporate a binary classifier and are updated based on a naive cross-entropy loss:
\begin{equation}
\label{classify loss}
{{\cal L}_{\mathrm{CE}}}\!=\!-\frac{1}{{2N}}\sum\limits_{j=1}^{2N}
{{y_j} \log ( {\rho( {{\mathbf f}_j} )})
\!+\!({1\!-\!{y_j}} ) \log ( {1\!-\!\rho( {{\mathbf f}_j} )} )},
\end{equation}
where $2N$ is the total number of samples in a mini-batch. $y_i$ is the ground-truth of feature ${\mathbf f}_i$. $\rho(\cdot)$ is the classifier that outputs the probability of belonging to the real and fake classes. 

Due to the inherent properties of binary cross-entropy loss, as $\rho({{\mathbf f}_j})$, the predicted probability of belonging to the fake class \emph{increases}, $1-\rho({{\mathbf f}_j})$, the probability of belonging to the real class will correspondingly \emph{decrease}, and vice versa. 
If the detector can abstract accurate criteria for identifying forgery faces, this `one rises as the other falls' single-decision logic will achieve a satisfactory optimization result. However, in real-world scenarios, the forgery features are often tangled with overfitted `mis-artifacts' and have poor generalization ability in the target domain. 
Therefore, we introduce a more reliable Independent Dual-Decision Classification~(IDC) strategy and produce the final classification probability based on both real facial concepts and forgery artifacts.

\noindent\textbf{Independent Dual-Decision Classifier.} As for a sample $\mathbf{f}_j$, to sever the one-to-one correspondence between the fake probability $\rho({{\mathbf f}_j})$ and the real probability $1-\rho({{\mathbf f}_j})$, we extend the output dimensions of the naive binary classifier to twice, and the classification process is as follows:
\begin{equation}
\mathbf{\hat f}_j = \hat{\rho} (\mathbf{f}_j) = {\rm Sofmax}( {\rm FC}( \mathbf{f}_j) ),
\end{equation}
where the Independent Dual-Decision Classifier $\hat{\rho}(\cdot)$ consists of the fully connected layer and the normalization layer. The output vector $\mathbf{\hat f}_j \in {\mathbb{R} ^{1 \times 4}}$ can be further represented as:
\begin{equation}
{{\mathbf{\hat f}_j} = \left\{ {\mathbf{\hat f}_j^0,\mathbf{\hat f}_j^1,\mathbf{\hat f}_j^2,\mathbf{\hat f}_j^3} \right\}}~
(s.t.\sum\limits_{m=0}^{3} {{{\mathbf{\hat f}}_j^m} = 1}),
\end{equation}
where the first and second dimensions of the output vector $\mathbf{\hat f}_j$ represent the probabilities of belonging to the real and fake classes, respectively. The third and fourth dimensions serve as the auxiliary components for our IDC strategy, whose function will be elaborated in the next section. In this setup, the constraints are imposed on all four dimensions, so that $\mathbf{\hat f}_j^0$ no longer entirely follows the variations of $\mathbf{\hat f}_j^1$.

\noindent\textbf{Independent Dual-Decision Regularization.}
After obtaining the relative outputs from $\hat{\rho}(\cdot)$, we try to further reduce the probabilities of belonging to the incorrect classes by appending an extra regularization term $\cal R$ to the classification loss:
\begin{equation}
{\cal R} = \left\{{\begin{array}{*{20}{l}}
{\sum\limits_{j=1}^{2N} {\beta \cdot d_j^2} }&{\mathrm{if}~|d| < 1} \\
{\sum\limits_{j=1}^{2N} {({|d_j| - \beta})} }&{\mathrm{otherwise}},
\end{array}}\right.
\end{equation}
where $2N$ and $y_j$ are consistent with that in Equation (\ref{classify loss}). $\beta$ is a predefined value to control the optimization intensity. ${d_j}$ is used to represent the discrepancy between the incorrect probabilities and their corresponding auxiliary components:
\begin{equation}
{d_j} = \left\{ {\begin{array}{*{20}{c}}
{\mathbf{\hat f}_j^1 - \mathbf{\hat f}_j^3} &{{y_j} = 0}\\
{\mathbf{\hat f}_j^0 - \mathbf{\hat f}_j^2} &{{y_j} = 1}.
\end{array}} \right.
\end{equation}
This way, the probability of belonging to the incorrect class will progressively approach the lower auxiliary component without being influenced by the correct probability. For instance, for a real sample ${\mathbf{\hat f}_j}$~(with the label of ${{y_j} = 0}$), $\cal R$ balances the value of $\mathbf{\hat f}_j^1$ and $\mathbf{\hat f}_j^3$ which consistently maintains a small value.
This means combining the classifier~$\hat{\rho}(\cdot)$ and regularization term $\cal R$, our Independent Dual-Decision strategy can rely on another optimization path~(\ie, leveraging the real facial concepts) to make robust judgments and implicitly reduces the reliance on forgery artifacts.

\subsection{Training Strategy}
Combining all modules, the network framework is optimized in an end-to-end manner based on the following loss function: 
\begin{equation}
\mathcal{L}_{\mathrm{total}} = \mathcal{L}_{\mathrm{CE}} + \lambda_1 \mathcal{L}_{\mathrm{Diversity}} + \lambda_2 \mathcal{L}_{\mathrm{Distinction}} + \lambda_3 \mathcal{R},
\label{eqn:loss}
\end{equation}
where $\mathcal{L}_{\mathrm{CE}}$ is a binary cross-entropy loss. $\mathcal{L}_{\mathrm{Diversity}}$ and $\mathcal{L}_{\mathrm{Distinction}}$ are used to learn comprehensive real prototypes. $\mathcal R$ is a regularization term for achieving the independent dual-decision strategy. \(\lambda_1\), \(\lambda_2\), and \(\lambda_3\) are the hyperparameters that balance the contribution of each term. Through this holistic loss function, the model optimizes classification results based on both comprehensive real concepts and forgery artifacts, thereby enhancing generalization in unseen target domains.
\section{Experiment}
\label{sec:expo}

\begin{table*}[t]
\centering
\scalebox{0.95}{
\begin{tabular}{l|rcccccc}
\toprule \midrule
\multirow{1}{*}{{Method}}         &\multicolumn{1}{c}{Venue}      
& \multicolumn{1}{|c}{Celeb-DF} & \multicolumn{1}{c}{DFD} & \multicolumn{1}{c}{DFDC} & \multicolumn{1}{c}{DFDCp} & \multicolumn{1}{c}{UADFV} & \multicolumn{1}{|c}{AVG} \\ \cmidrule(lr){1-1} \cmidrule(lr){2-2} \cmidrule(lr){3-3} \cmidrule(lr){4-4} \cmidrule(lr){5-5} \cmidrule(lr){6-6} \cmidrule(lr){7-7} \cmidrule(lr){8-8} 
    
\multicolumn{1}{l|}{$^{\ddag}$EfficientNet~\cite{Efficient}}      & ICML'19  
& \multicolumn{1}{|c}{64.59}  & 92.31   &  65.43   &  80.27 & 63.19  & \multicolumn{1}{|c}{73.16}       \\
\multicolumn{1}{l|}{$^{\ddag}$Face X-ray~\cite{x-ray}}                     & CVPR'20    
& \multicolumn{1}{|c}{74.76} & 93.47  &  61.57   & 71.15  & 64.34  & \multicolumn{1}{|c}{73.06}       \\
\multicolumn{1}{l|}{$^{\ddag}$CORE~\cite{CORE_CVPRW_2022}}   &CVPRW'22   
& \multicolumn{1}{|c}{79.45}  & 93.74  &  62.60   & 75.74  &  65.41 & \multicolumn{1}{|c}{75.39}       \\
\multicolumn{1}{l|}{$^{\ddag}$RECCE~\cite{Face_Reconstruction}}   &CVPR'22  
& \multicolumn{1}{|c}{69.71} & 93.15  & 62.82  & 74.19  & 78.61  & \multicolumn{1}{|c}{75.70}       \\
\multicolumn{1}{l|}{$^{\ddag}$SBI~\cite{SBI_ShioharaY22}}         & CVPR'22            
& \multicolumn{1}{|c}{93.18} &97.56 &72.42 &86.15 &97.28 & \multicolumn{1}{|c}{89.32}       \\
\multicolumn{1}{l|}{$^{\ddag}$UCF~\cite{UCF_0002ZFW23}}           & ICCV'23          
& \multicolumn{1}{|c}{81.90} & 93.09  & 66.21  &  80.94 & 97.15  & \multicolumn{1}{|c}{83.86}       \\
\multicolumn{1}{l|}{FoCus~\cite{tian2024learning}}      & TIFS'24          
& \multicolumn{1}{|c}{76.13}&-    &68.42&76.62& -  & \multicolumn{1}{|c}{-}       \\
\multicolumn{1}{l|}{Qiao et al.~\cite{1}}   & TPAMI'24          
& \multicolumn{1}{|c}{70.00}&94.00&-    &-    &78.00& \multicolumn{1}{|c}{-}       \\
\multicolumn{1}{l|}{GRU~\cite{2}}      & CVPR'24          
& \multicolumn{1}{|c}{89.00}  &96.10  &  -   & -  & -  & \multicolumn{1}{|c}{-}       \\
\multicolumn{1}{l|}{ProDet~\cite{cheng2024leavedeepfakedatatraining}}             &   NeurIPS’24
& \multicolumn{1}{|c}{84.48}  & -  &  72.40  & 81.16  & -  & \multicolumn{1}{|c}{-}       \\
\midrule  
\rowcolor[HTML]{D9EEF2}\multicolumn{1}{l|}{RealID}      & \multicolumn{1}{c}{-}   
& \multicolumn{1}{|c}{\textbf{95.16}} & \textbf{98.32} &\textbf{74.67} &\textbf{88.80}  &\textbf{98.34}  & \multicolumn{1}{|c}{\textbf{91.06}}       \\
\midrule 
\bottomrule
\end{tabular}}
\caption{Performance comparison (\%). All models are trained on the FF++ dataset. The best performance is marked as bold. $^{\ddag}$: We re-implemented this detector. -: The authors did not report the results on this dataset in their original paper.}
\label{Tab:SOTA}
\end{table*}

\subsection{Implementation}
We utilized Dlib\footnote{http://dlib.net/.} to extract faces and resize them to 256 $\times$ 256 pixels for both the training and testing sets. We conducted the experiments on a single RTX 3090 GPU with a batch size of 16. The backbone we employed is EfficientNet~\cite{Efficient} and we also switched the backbone to other networks, \textit{e.g.}, ViT~\cite{ViT}, to demonstrate the robustness of our proposed method in Section~\ref{sec:exp_backbone}.
The hyperparameters $\lambda_1$, $\lambda_2$, and $\lambda_3$ in Equation~(\ref{eqn:loss}) are selected via grid search and set to 0.6, 1.0, and 1.0, respectively. 
Similar to the common setup for generalizable deepfake detection~\cite{RFM,Learning_Second_Order,Face_Reconstruction}, we trained our model with the FF++ dataset~\cite{Xception}. This dataset includes 1,000 real videos from YouTube, as well as five types of manipulated videos yielding a total of 6,000 videos. Finally, to evaluate the generalization capability of our model, we performed the cross-dataset testing on five widely used deepfake datasets, \ie, 
Celeb-DF~\cite{Celeb-DF},
DFD~\cite{google_deepfake_dataset},
DFDC~\cite{dfdc},
DFDCp~\cite{dfdc}, and
UADFV~\cite{li2018ictu}.

\subsection{Main Experimental Results}
\textbf{Performance on Cross-Dataset Evaluation.} We reported the generalization capability of several SoTA baselines and our RealID in Table~\ref{Tab:SOTA}. All of them are trained on FF++ and evaluated on other five testing datasets. This cross-dataset setup is challenging since neither the testing pristine/forged videos nor the manipulated techniques are visible in the training dataset. 
We utilized the AUC (area under the receiver operating characteristic curve) metric to quantify the performance. In addition to reporting the AUCs on each individual dataset, we also calculated the average AUC over the five testing datasets. 
From Table~\ref{Tab:SOTA}, we have three main observations:
1)~Our method RealID significantly improves the generalization ability compared to several SoTA baselines. For instance, our method achieves an AUC improvement of nearly 20\% on the Celeb-DF dataset compared to FoCus~\cite{tian2024learning} and an average AUC improvement of approximately 18\% across five datasets compared to EfficientNet.
2)~Our method is more robust. Most of the baselines are only effective on specific datasets, whereas our method achieves significant improvements across all datasets. For instance, RECCE, which demonstrates competitive performance on other datasets, unexpectedly suffers a significant performance drop on Celeb-DF. In contrast, our RealID approach consistently achieves effective improvements.
3)~Exploring detailed image features can enhance performance. For instance, training with self-blended images enables SBI to achieve an impressive average AUC of 89\%, and UCF leverages the common forgery features to achieve an AUC of 97\% on the UAVDF dataset, outperforming other baselines.

\label{sec:exp_backbone}
\begin{table}[t]
\centering
\scalebox{0.8}{
\begin{tabular}{l|cccccc}
\toprule \midrule
\multirow{1}{*}{Backbone}           
& \multicolumn{1}{|c}{Celeb-DF} & \multicolumn{1}{c}{DFDC} & \multicolumn{1}{c}{DFDCp} & \multicolumn{1}{c}{UADFV} &\multicolumn{1}{c}{AVG} \\ \cmidrule(lr){1-1} \cmidrule(lr){2-2} \cmidrule(lr){3-3} \cmidrule(lr){4-4} \cmidrule(lr){5-5} \cmidrule(lr){6-6}

\multicolumn{1}{l|}{Xception}               
& \multicolumn{1}{|c}{56.75}  &  64.19   &  74.17  & 62.05  &64.29  \\
\multicolumn{1}{l|}{Xception+RealID}               
& \multicolumn{1}{|c}{87.44}  &  73.21   & 81.89   & 99.23  &85.44   \\ \midrule
\multicolumn{1}{l|}{ViT-L}                         
& \multicolumn{1}{|c}{77.27}  &  71.83   & 84.05   & 76.13  &77.32 \\
\multicolumn{1}{l|}{ViT-L+RealID}                         
& \multicolumn{1}{|c}{90.00}  &  81.80	 & 91.60   & 98.92  &90.77 \\ \midrule
\multicolumn{1}{l|}{ViT-B}                          
& \multicolumn{1}{|c}{89.63}  &  73.00   & 87.04   & 96.17  &86.46  \\
\multicolumn{1}{l|}{ViT-B+RealID}                          
& \multicolumn{1}{|c}{93.56}  &  80.92	 & 92.30   & 98.85  &91.40\\
\midrule 
\bottomrule
\end{tabular}}
\caption{Validate the effectiveness of RealID on different backbones. ViT-L and Vit-B represent different scales of the ViT.}
\label{Tab:backbone}
\end{table}
\noindent\textbf{Performance with Different Backbones.}
By switching the backbone, we further demonstrated the generalizability of our method.
Specifically, we selected three widely employed backbones: Xception~\cite{Xception}, ViT-L-32~\cite{ViT}, and ViT-B-16~\cite{ViT} to evaluate the performance of combining them with our RealID. The training details remain consistent with those that utilize EfficientNet. It is worth noting that RealID can be seamlessly integrated into any deepfake backbone since it does not require modifications to the original architecture.

We have reported the experimental results in Table~\ref{Tab:backbone}. From this table, we have the following observations: 
1)~Our RealID remains effective across different backbones. For instance, Xception+RealID achieves a 21.15\% improvement in average AUC compared to the Xception alone, while ViT-L+RealID and ViT-B+RealID achieve 12.73\% and 3.93\% improvements on the Celeb-DF dataset, respectively. The above results fully demonstrate the generalization of our approach.
2)~Stronger backbone architectures yield better performance but are harder to improve. For instance, the ViT-B-based model outperforms both EfficientNet and Xception in terms of average AUC. However, the performance improvement ratio of ViT-B+RealID is the smallest. 
3)~Appropriate patch size is more important than the model scale. For instance, ViT-B-16 with a patch size of 16 outperforms ViT-L-32 with a patch size of 32. This may also indicates the importance of local features in deepfake detection.

\subsection{Ablation Studies}
\begin{table}[t]
\centering
\scalebox{0.75}{
\begin{tabular}{ccc|cccc}
\toprule
\midrule
\multirow{2}{*}{Backbone} & \multirow{2}{*}{$\mathrm{RealC^2}$}   &\multirow{2}{*}{$\mathrm{IDC}$}    
&\multicolumn{4}{c}{Testing Set}                   \\ \cmidrule{4-7} 
& & &Celeb-DF &DFDC &DFDCp &UADFV                                \\ \midrule
\Checkmark  & &  
&{64.59} &{65.43} &{80.27} &{63.19} \\
\Checkmark &\Checkmark  & 
&{94.92}&{73.58}&{88.11}&{98.09} \\ 
\Checkmark &  &\Checkmark
&{93.79}&{74.64}&{88.44}&{98.30} \\ 
\Checkmark &\Checkmark &  \Checkmark
&\textbf{95.16} &\textbf{74.67} &\textbf{88.80}  &\textbf{98.34} \\
 \midrule \bottomrule
\end{tabular}
}
\caption{AUC (\%) comparison of different modules in RealID.}
\label{tab:ablation}
\end{table}
\noindent\textbf{Module Analysis.} In Table~\ref{tab:ablation}, we presented the results of ablation studies on two key modules in RealID, \ie, $\mathrm{RealC^2}$ and $\mathrm{IDC}$. From the results, it can be observed that both components effectively enhance the generalizability of the model. For instance, compared with the backbone on the Celeb-DF dataset, $\mathrm{RealC^2}$ and $\mathrm{IDC}$ improve the AUC by 30\% and 29\%, respectively. Combining the two components, the best result, a 30.57\% AUC improvement, can be achieved. 

\begin{table}[t]
\centering
\scalebox{0.80}{
\begin{tabular}{cc|cccc}
\toprule
\midrule
\multirow{2}{*}{$\mathcal{L}_{\mathrm{Diversity}}$} &\multirow{2}{*}{$\mathcal{L}_{\mathrm{Distinction}}$}      
&\multicolumn{4}{c}{Testing Set}                   \\ \cmidrule{3-6} 
 & & Celeb-DF&DFDC&DFDCp&UADFV                                \\ \midrule
\Checkmark & 
& 91.83 & 72.22 & 75.37 & 95.60\\
 &\Checkmark
&{93.87}&{72.49}&{87.81}&{97.70} \\
 \midrule \bottomrule
\end{tabular}
}
\caption{AUC (\%) comparison of loss functions in $\mathrm{RealC^2}$.}
\label{tab:ablation_loss}
\end{table}
\noindent\textbf{Loss Analysis.} To ensure the diversity and distinction of the real prototypes used in $\mathrm{RealC^2}$, we employ two relative loss functions, \ie,  $\mathcal{L}_{\mathrm{Diversity}}$ and $\mathcal{L}_{\mathrm{Distinction}}$, to supervise the training process. 
Specifically, $\mathcal{L}_{\mathrm{Distinction}}$ ensures the consistency of features within a subclass while maintaining the separability between the real and fake classes. $\mathcal{L}_{\mathrm{Diversity}}$ enforces diversity among prototypes, preventing them from converging into the same distribution.
In Table~\ref{tab:ablation_loss}, we reported the impacts of these two loss functions. It can be seen that, between the two loss functions, $\mathcal{L}_{\mathrm{Distinction}}$ plays a more significant role. For instance, the AUC on the Celeb-DF dataset will drop by 3.3\% when $\mathcal{L}_{\mathrm{Distinction}}$ is removed. 
One critical reason for this is that $\mathcal{L}_{\mathrm{Distinction}}$ updates the real prototypes based on real instances. Without it, the real prototypes cannot undergo global optimization.
Furthermore, $\mathcal{L}_{\mathrm{Diversity}}$ is also crucial, contributing approximately a 1.2\% improvement. In summary, the combination of these two loss functions in $\mathrm{RealC^2}$ delivers the best performance gains.

\noindent\textbf{Parameter Analysis.}
\label{sec:parameter}
\begin{figure}[t]
    \centering
    \includegraphics[width=0.45\textwidth]{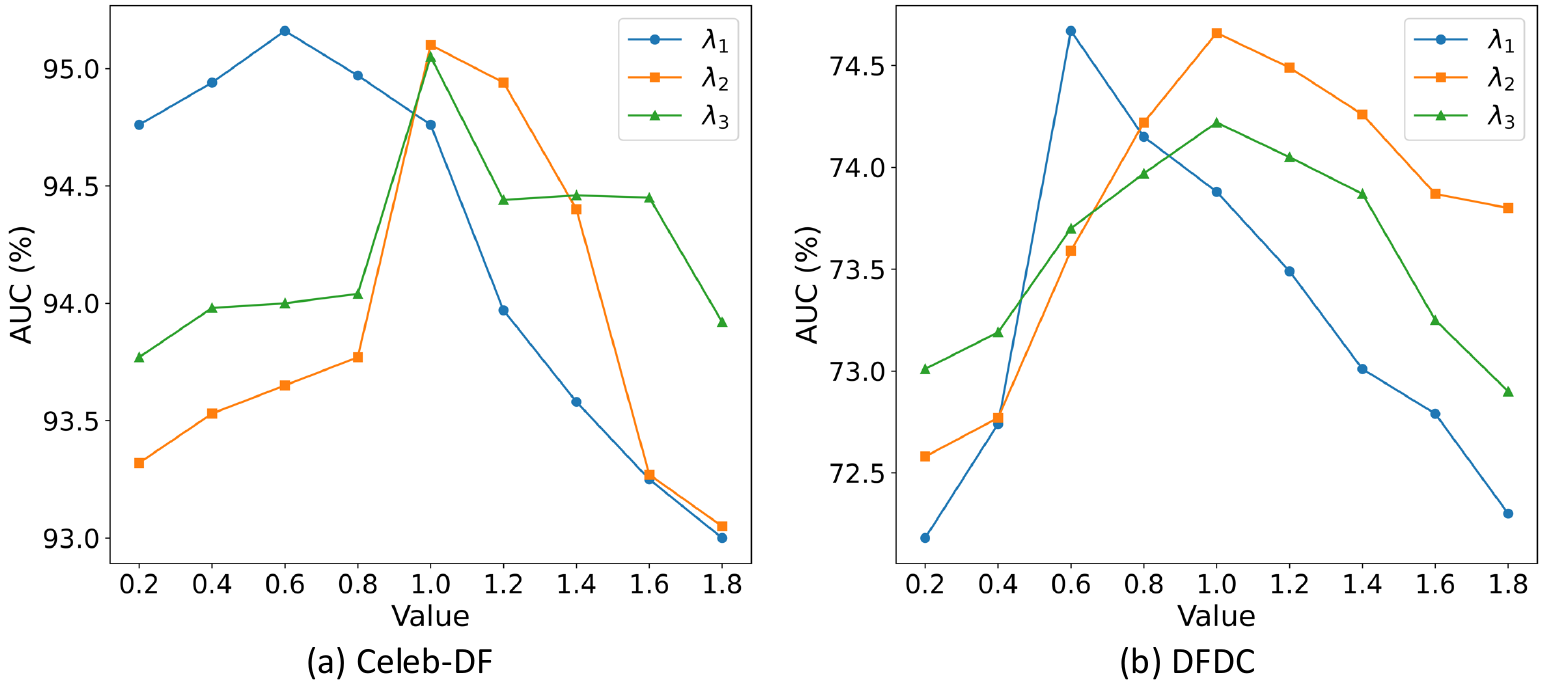}
    \caption{AUC (\%) comparison under different hyperparameter combinations. For $\lambda_1$, $\lambda_1$, and $\lambda_1$, we vary one of their values while keeping the other two values fixed at 0.5.}
    \label{fig:para_analysis}
\end{figure}
In Equation~(\ref{eqn:loss}), the three parameters, \ie, $\lambda_1$, $\lambda_2$, and $\lambda_3$, control the influence of the loss functions. Figure~\ref{fig:para_analysis} illustrates the impacts of these parameters on the performance. Specifically, we progressively increase a certain hyperparameter within the range of [0, 2], while keeping the other two parameters fixed, to retrain the model. The performance is then evaluated on DFDC and Celeb-DF.
From the figure, we observed that the impact of the three hyperparameters on the model's performance on both datasets follows the same pattern: as the corresponding $\lambda$ value increases, the model's performance first improves and then drops sharply. Based on the combined performance of the three parameters, we empirically selected $\lambda_1 = 0.6$, $\lambda_2 = 1.0$, and $\lambda_3 = 1.0$ as the optimal configuration.

\subsection{Qualititive Studies}
\begin{figure}
    \centering
    \includegraphics[width=0.45\textwidth]{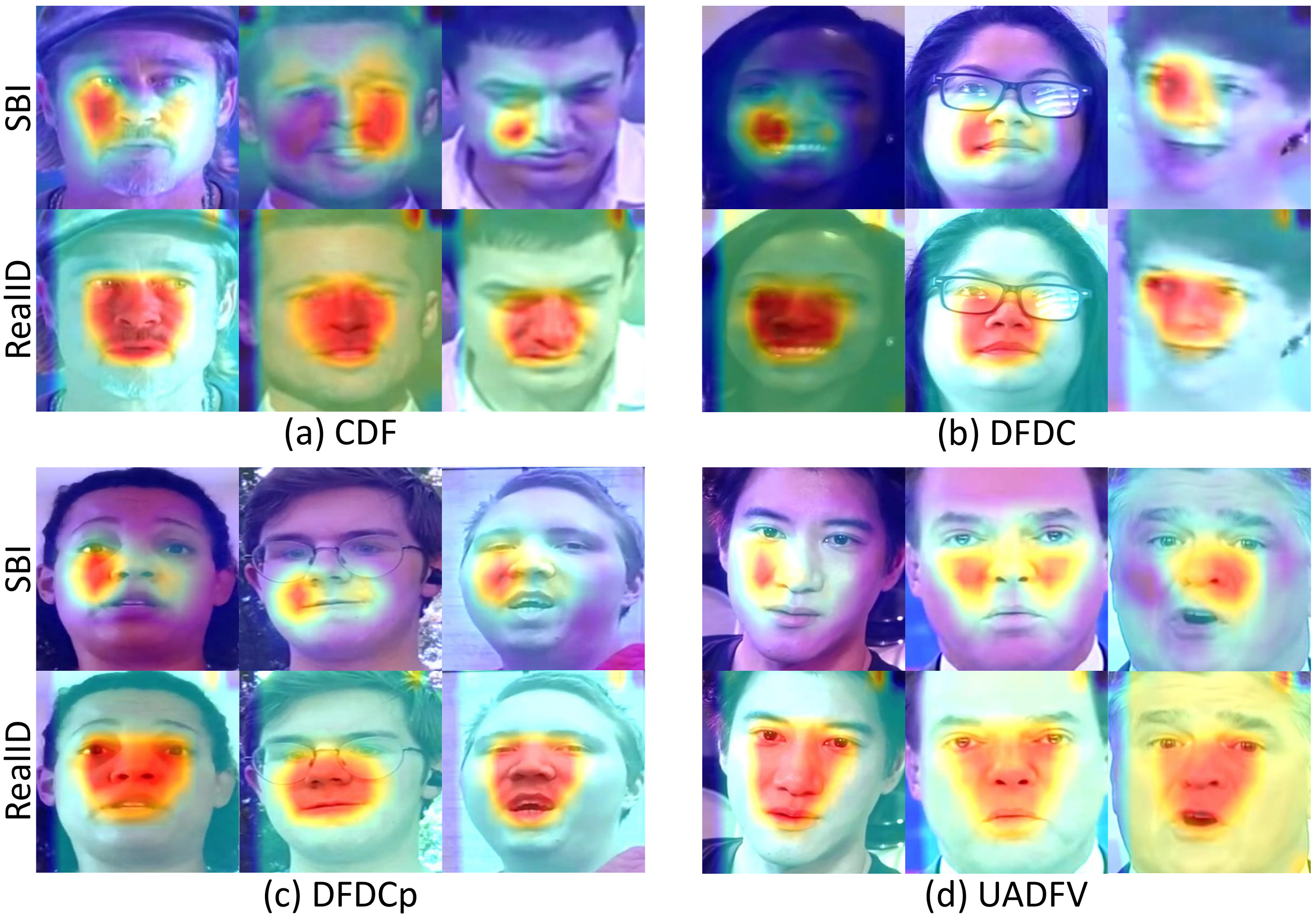}
    \caption{Illustration for no-cherry-pick heatmaps from four datasets, comparing our RealID with the SoTA baseline SBI.}
    \label{fig:heatmaps}
\end{figure}
In Figure~\ref{fig:heatmaps}, we presented heatmaps of no-cherry-pick instances from different datasets. It can be observed that, compared to SBI~\cite{SBI_ShioharaY22}, the baseline model, our approach achieves a more evenly distributed attention across the entire face. For example, on the DFDC dataset, SBI focuses primarily on the left cheek, whereas our approach distributes attention more evenly across the lips, nose, and eyes. This indicates that our RealID makes decisions based on more comprehensive facial concepts rather than the presence or absence of local forgery traces.

\begin{figure}[t]
\centering
\includegraphics[width=0.42\textwidth]{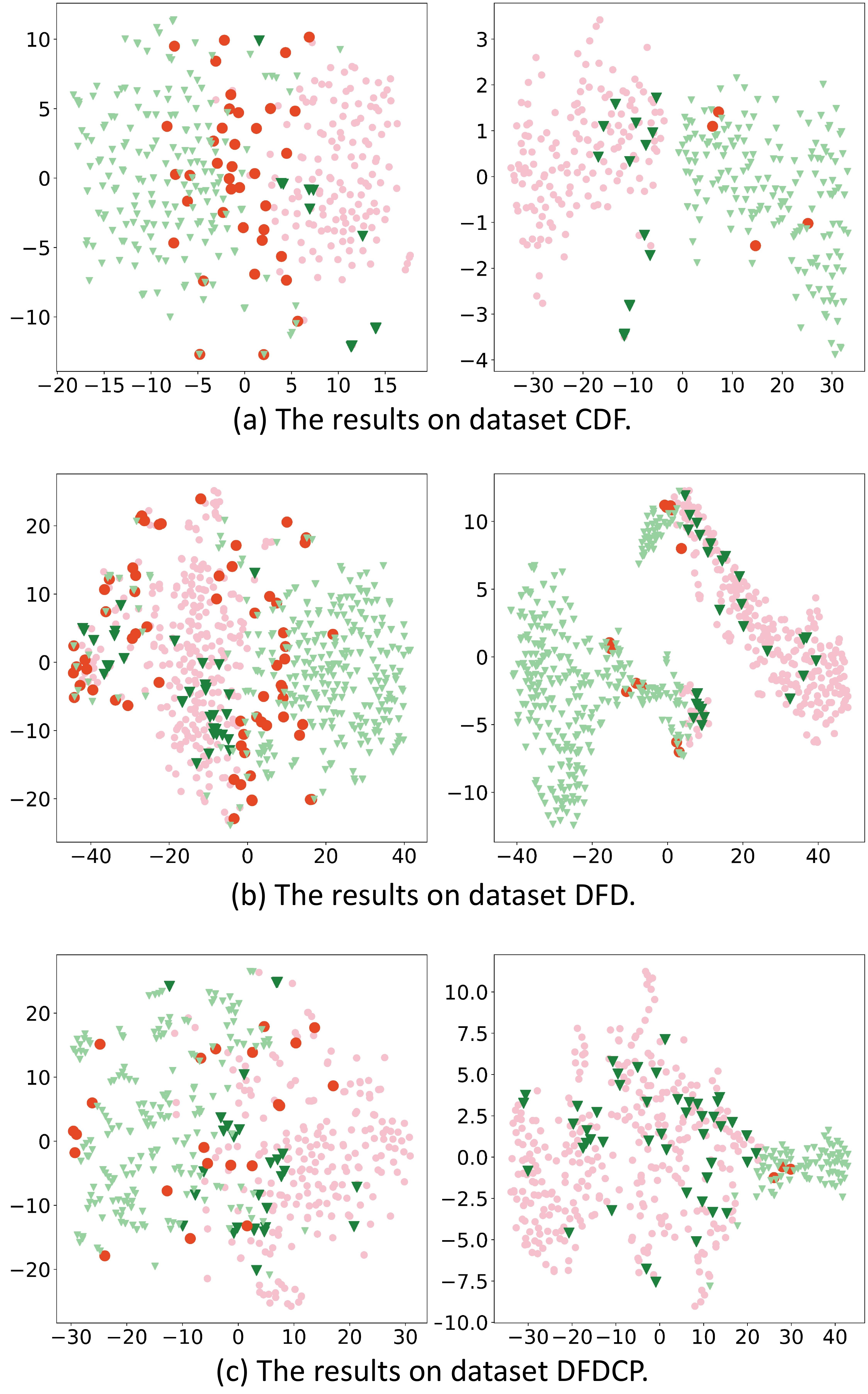}
\caption{t-SNE visualization. \textbf{Left:} The features extracted by the baseline model. \textbf{Right:} The features extracted by our RealID model.}
\label{fig:tsne-2}
\end{figure}

In Figure~\ref{fig:tsne-2}, we reported the t-SNE visualization for features in testing datasets. For each subfigure, the left side represents the distribution of test samples under SBI, while the right side shows the distribution under our RealID. We can observe that our RealID significantly reduces the misclassified real samples (marked as large red circles). For instance, a notable number of outliers are corrected on the CDF dataset. 

\section{Conclusion} 
In this work, we address the critical challenge of generalization obstacle of deepfake detection models, particularly their tendency to misclassify real instances as fake when applied to unseen datasets. We identify that this issue stems from an over-reliance on forgery artifacts and a limited understanding of `real'. To overcome these limitations, we propose RealID, a novel approach comprising the Real Concept Capture Module and the Independent Decision Classification Module. 
Extensive experiments demonstrated that RealID significantly outperforms state-of-the-art baselines in cross-dataset detection scenarios. It not only advances the field of deepfake detection but also offers a solution for future research on balanced and robust detection strategies.



\bibliographystyle{named}
\bibliography{ijcai25}

\end{document}